\documentclass[journal]{IEEEtran}
\usepackage{url}
\usepackage{cite}
\usepackage{xcolor}
\usepackage{textcomp}
\usepackage{graphicx}
\usepackage{listings}
\usepackage{subcaption}
\usepackage{algorithmic}
\usepackage[hidelinks]{hyperref}
\usepackage{amsmath,amssymb,amsfonts}

\graphicspath{{.}}
\definecolor{CodeColor}{HTML}{0000c0}
\newcommand{\cinline}[1]{\lstinline[columns=fullflexible]{#1}}

\definecolor{LinkColor}{HTML}{c00000}
\hypersetup{colorlinks=true, citecolor=LinkColor,
            linkcolor=LinkColor, urlcolor=LinkColor}

\def\BibTeX{{\rm B\kern-.05em{\sc i\kern-.025em b}\kern-.08em
    T\kern-.1667em\lower.7ex\hbox{E}\kern-.125emX}}

\begin{document}

\title{Spyker: High-performance Library for Spiking Deep Neural Networks}

\author{
\IEEEauthorblockN{
Shahriar Rezghi Shirsavar\IEEEauthorrefmark{2}\IEEEauthorrefmark{3},
Mohammad-Reza A. Dehaqani\IEEEauthorrefmark{2}\IEEEauthorrefmark{3},
}

\IEEEauthorblockA{\IEEEauthorrefmark{2}School of Electrical and Computer Engineering, College of Engineering, University of Tehran, Tehran, Iran\\\{shahriar.rezghi, dehaqani\}@ut.ac.ir}

\IEEEauthorblockA{\IEEEauthorrefmark{3}School of Cognitive Sciences, Institute for Research in Fundamental Sciences (IPM), Tehran, Iran}

\IEEEauthorblockA{\IEEEauthorrefmark{1}Corresponding author: Mohammad-Reza A. Dehaqani, dehaqani@ut.ac.ir}
}

\maketitle

\begin{abstract}

Spiking neural networks (SNNs) have been recently brought to light due to their promising capabilities. SNNs simulate the brain with higher biological plausibility compared to previous generations of neural networks. Learning with fewer samples and consuming less power are among the key features of these networks. However, the theoretical advantages of SNNs have not been seen in practice due to the slowness of simulation tools and the impracticality of the proposed network structures. In this work, we implement a high-performance library named Spyker using C++/CUDA from scratch that outperforms its predecessor. Several SNNs are implemented in this work with different learning rules (spike-timing-dependent plasticity and reinforcement learning) using Spyker that achieve significantly better runtimes, to prove the practicality of the library in the simulation of large-scale networks. To our knowledge, no such tools have been developed to simulate large-scale spiking neural networks with high performance using a modular structure. Furthermore, a comparison of the represented stimuli extracted from Spyker to recorded electrophysiology data is performed to demonstrate the applicability of SNNs in describing the underlying neural mechanisms of the brain functions. The aim of this library is to take a significant step toward uncovering the true potential of the brain computations using SNNs.

\end{abstract}

\begin{IEEEkeywords}
Spiking Neural Network, Learning Rules, C++/CUDA, Modular Structure, Biological Plausibility
\end{IEEEkeywords}

\section{Introduction}

The human brain can operate with amazing robustness and energy efficiency. Artificial neural networks (ANNs) aim at modeling the brain, and three generations of these networks have been developed. Each generation of ANNs improves the quality of the modeling of the brain compared to the last. The first generation of ANNs makes use of the McCulloch-Pitts neurons \cite{mcculloch_logical_1943}. Although these neurons are inspired by biological neurons, time dynamics are not considered in this model, and the learning rules proposed for them lack power and biological plausibility. These neurons were used in multi-layer perceptron (MLPs) \cite{rosenblatt_perceptron_1958} and Hopfield \cite{hopfield_neural_1982} networks.

The second generation of ANNs uses a continuous activation function (ReLU \cite{nair_rectified_2010} and sigmoid \cite{rumelhart_learning_1986}, for example) instead of thresholding, which makes them suitable for processing analog signals. They have attracted the attention of researchers in recent years and were able to reach high accuracies \cite{graves_framewise_2005, vaswani_attention_2017} (even surpassing humans) and win different challenges \cite{zhai_scaling_2022}. Despite the success of DNNs, there are structural differences between these networks and the human brain. Lack of temporal dynamics, using analog signals for network propagation and activation functions, learning rules without biological roots, and the need for large amounts of data \cite{sun_revisiting_2017} and energy \cite{li_evaluating_2016} to achieve acceptable results are among these differences.

The third generation of neural networks is spiking neural networks (SNNs). The neural models used in these networks simulate biological neurons more accurately, and the coding mechanisms used in these networks are found in neural communications. Furthermore, the learning rules used in these networks have been discovered in the brain \cite{mcmahon_stimulus_2012, meliza_receptive-field_2006, huang_associative_2014}. Having lower energy consumption, learning with fewer samples, and solving more complicated tasks due to time dynamics (several electrophysiological studies emphasize the role of temporal dynamics in neural coding \cite{dehaqani_temporal_2016, dehaqani_selective_2018}) are some of the advantages of SNNs compared to the second generation of ANNs. SNNs can be used to solve machine learning tasks, study and explore brain functionality, and run on specialized hardware with low power consumption. The research being done on these networks aims to address the disadvantages of DNNs with more realistic modeling of the brain functionality.

\begin{figure*}[tb]
\centerline{\includegraphics[width=\textwidth]{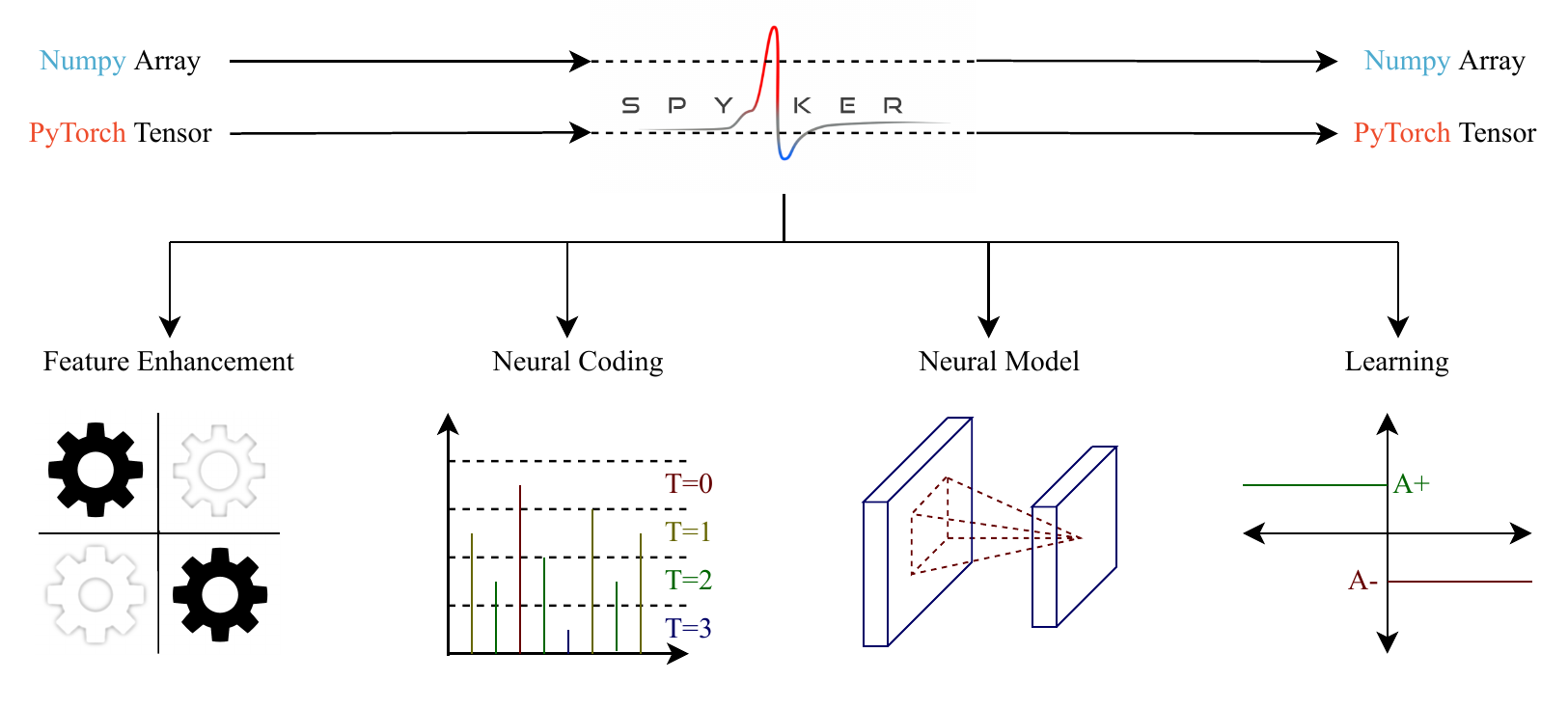}}
\caption{Overview of the Spyker library. Spyker API supports PyTorch tensors and Numpy arrays as well as a built-in data wrapper. The output of Spyker operations have the same container type as the input. The functionality of Spyker can be grouped into subcategories shown in the figure.}
\label{fig:library-overview}
\end{figure*}

Several high-performance well-established frameworks like PyTorch \cite{paszke_pytorch_2019}, TensorFlow \cite{developers_tensorflow_2022}, and MXNet \cite{chen_mxnet_2015} have been developed for DNNs in recent years. These libraries have enabled DNNs to achieve new highs in solving machine learning tasks. SNNs are not yet comparable to DNNs due to the lack of fast simulation tools. There have been some attempts, like SpykeTorch \cite{mozafari_spyketorch_2019} and BindsNet \cite{hazan_bindsnet_2018}. SpykeTorch, written on top of the PyTorch framework, is a simulator for large-scale spiking neural networks (SDNNs). However, it has a slow runtime, and training even simple networks can take up to days to complete. To our knowledge, Spyker is the first toolbox to simulate large-scale networks with high performance, is easy to use, has the flexibility to be used in multiple languages, and has the compatibility to integrate with other commonly used tools. In order to fill this need, we have developed Spyker. Spyker is a C++/CUDA library written from scratch with both C++ and Python interfaces and support for dense and sparse structures. Although Spyker is a stand-alone library, it has a highly flexible API and can work with PyTorch tensors and Numpy arrays. Figure \ref{fig:library-overview} shows an overview of the library. In order to increase performance, small-sized integers are used alongside floating-point numbers. It also uses highly-optimized low-level back-end libraries such as OneDNN and cuDNN to speed up heavy computations such as convolutions and matrix multiplications. Spyker can be compiled on various CPUs to be optimized locally and take advantage of native CPU-specific instructions.

Spiking neural networks are made of different building blocks (see \cite{shirsavar_models_2022} for more details). The first block is the modeling of the biological neurons. Some examples of this are leaky integrate-and-fire \cite{abbott_lapicques_1999}, spike-response model \cite{jolivet_spike_2003}, and Izhikevich model \cite{izhikevich_simple_2003}. Another building block is neural coding, which can be rate coding \cite{adrian_impulses_1926}, temporal coding, phase coding and synchrony coding \cite{buzsaki_rhythms_2006}, or other coding schemes. The final building block is the learning mechanism. Examples of these mechanisms are STDP \cite{markram_regulation_1997, gerstner_neuronal_1996}, R-STDP \cite{fremaux_neuromodulated_2016}, backpropagation \cite{bohte_spikeprop_2000}, and conversion from ANNs to SNNs \cite{diehl_fast-classifying_2015}. Spyker has a modular implementation of these three blocks that enables its users to build SNNs.

Spyker provides SNN functionality with a high-performance and easy-to-use interface with an open-source and permissive license. It can run on CPU and CUDA devices and has both dense and sparse interfaces. The library introduces new features and fixes most of the shortcomings of its predecessor. The improvements include adding batch processing, strided convolutions, internal padding for convolutions, fully connected layers, and the rate coding mechanism. Compared to its predecessor, the interface of the library is simpler, closer to the current API of deep learning libraries, and more straightforward to use. In this work, several successful network structures are implemented using this library to prove its operability, its runtime speed is compared to SpykeTorch, and the results indicate Spyker can run up to eight times faster. The proposed work is able to reduce the gap between SNNs and DNNs and bring us a step closer to uncovering the true potential of spiking neural networks.

We start with a description of dimensionality of the input arrays and how the spike trains are implemented in the library. Afterward, we provide an explanation of different building blocks of SNNs and how they are implemented in Spyker and modeled in the interface. Then, we implement network structures that have been succesful to prove its operatibility, and we compare the performance of the library to its predecessor on these networks. Furthermore, comparison of the represented stimuli extracted from Spyker to recorded electrophisiology data is performed to demonstrate the applicability of SNNs in describing the underlying neural mechanisms of the brain functions. Finally, we demonstrate an example usage of the library and discuss the impacts of this work and how it can be further improved.

%%%%%%%%%%%%%%%%%%%%%%%%%%%%%%%%%%%%%%%%%%%%%%%%%%%%%%%%%%%%%%%%%%%%%%%%%%%%%%%%
% METHODS %%%%%%%%%%%%%%%%%%%%%%%%%%%%%%%%%%%%%%%%%%%%%%%%%%%%%%%%%%%%%%%%%%%%%%
%%%%%%%%%%%%%%%%%%%%%%%%%%%%%%%%%%%%%%%%%%%%%%%%%%%%%%%%%%%%%%%%%%%%%%%%%%%%%%%%

\section{Methods}

The interface of the Spyker can be better explained when the classes and methods of the interface are grouped by building blocks of SNNs. The categories are feature enhancement, neural coding, neural model, and learning. In this section, the structure of the input to the network is explained. Afterward, the sparse and the dense interfaces are compared. Finally, the building blocks of the library are discussed in detail.

\subsection{Network Input}

Arrays passed through convolutional neural networks that process images are often four-dimensional arrays composed of batch size (B or N), number of channels (C), image height (H), and image width (W). The order can either be BCHW or BHWC  (or NCHW or NHWC). SNNs have temporal dynamics, and it is implemented as a dimension that represents time steps in Spyker. The library implements five-dimensional arrays with BTCHW order (T being the time steps). Since DNNs process analog signals, data types used in these networks are (usually four-byte) floating-point numbers. This data type can be computationally expensive compared to a small-sized integer type and take up more space in the memory. Since SNNs process binary signals, Spyker can optionally use eight-bit (or wider) integers alongside floating-point numbers to improve performance further.

\subsection{Dense vs Sparse interface}

The dense interface of Spyker uses the fully allocated memory buffers that are used in neural network computations. However, the sparse interface only needs to hold the indices of the spikes. Conversion between dense and sparse interfaces are provided in the library. The sparse interface has some advantages compared to the dense interface. In the dense interface, the time consumed by each operation is a function of the size of each of the 5 dimensions. However, in the sparse interface, it depends on the number of spikes. This means both memory and time consumed will be greatly reduced when processing sparser signals. Furthermore, since neurons fire at most once when using rank order coding, the increment of the number of time steps will have a smaller effect in the sparse interface compared to the dense interface.

\subsection{Feature Enhancement}

A transformation can be used to enhance features of the input signal (image) before the neural coding process \cite{kheradpisheh_stdp-based_2018, mozafari_bio-inspired_2019, falez_improving_2020}. This results in highlighted features having higher intensities and appearing in earlier time steps, meaning more excitation. Feature enhancement is done through filtering the input here. Various filters are supported in Spyker, and they are introduced in the following subsections.

\subsubsection{Difference of Gaussian Filter}

The first filter is the Difference of Gaussian (DoG). This filter increases the intensities of edges and other details in the image (see Figure \ref{fig:filter-example} for an example) \cite{el-sennary_edge_2019}. It approximates the center-surround properties of the ganglion cells of the retina \cite{rodieck_quantitative_1965} (see also \cite{bonin_suppressive_2005, lindeberg_computational_2013}). This operation is implemented as \cinline{spyker.DoG(size, filters, pad, device)} where \cinline{size} is the size of the width and the height of the filter, \cinline{filters} is a list of DoG filter descriptions (each description takes in two standard deviations), \cinline{pad} is the size of the padding of the image, and \cinline{device} is the device the filter will run on (CPU, GPU or others).

\subsubsection{Gabor Filter}

The following filter is the Gabor filter that determines the presence of specific frequency in content in a specific direction in the image. Research Indicates \cite{olshausen_emergence_1996} that the Gabor filter is used in the human visual cortex. The Gabor filter is implemented as \cinline{spyker.Gabor(size, filters, pad, device)}. The parameters of this class are the same as the DoG class, but the \cinline{filters} are Gabor filter descriptions, and each description takes in sigma, theta, gamma, lambda, and psi.

\subsubsection{Laplacian of Gaussian Filter}

The Laplacian of Gaussian (LoG) layer is also implemented in Spyker, and it is approximated using two DoG filters. An LoG filter with standard deviation $\sigma$ can be approximated using two DoG filters with ($\sigma \sqrt 2$, $\sigma / \sqrt 2$) and ($\sigma / \sqrt 2$, $\sigma \sqrt 2$) standard deviations. This filter exists in Spyker as \cinline{spyker.LoG(size, stds, pad, device)} where \cinline{stds} are a list of standard deviations needed to describe multiple LoG filters.

\begin{figure*}[tb]
\centerline{\includegraphics[width=\textwidth]{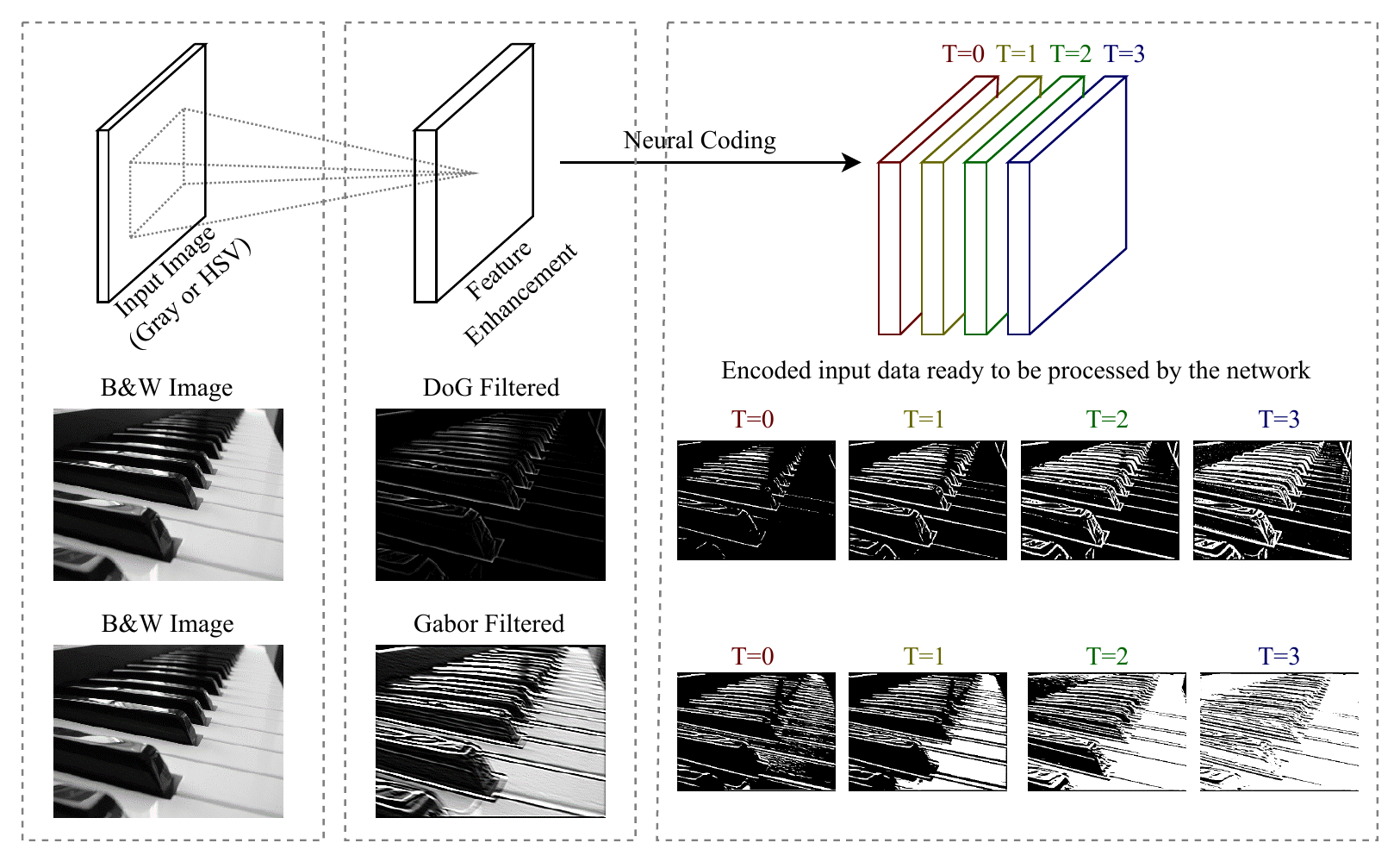}}
\caption{The figure shows a black and white image being filtered by DoG and Gabor filters. The theta parameter of the Gabor filter is set to -15 degrees. Then the images are coded using rank order coding into four time steps. Spikes are shown with white color on a black background through time steps. Spikes carry on from the previous to the current time step (cumulative structure).}
\label{fig:filter-example}
\end{figure*}

\subsubsection{Shape of the Filters}
The previously explained filters have kernel size $K_c \times K_h \times K_w$, which are square kernels ($K_h = K_w$). The input can have $B \times C_i \times H_i \times W_i$ shape which corresponds to batch, channels, height, and width of the input, respectively. The output will have $B \times C_o \times H_o \times W_o$ shape where:
\begin{equation}
\begin{aligned}
C_o &= C_i \times K_c \\
H_o &= H_i + 2 \times P_h - K_h + 1 \\
W_o &= W_i + 2 \times P_h - K_w + 1
\end{aligned}
\end{equation}
and $P_h$ and $P_w$ are height and width padding of the filter. The $K_c$ filters are applied to each channel separately.

\subsubsection{Zero-phase Component Analysis}

Final implemented layer is zero-phase component analysis (ZCA) Whitening. It has been suggested \cite{falez_improving_2020} that this transformation can improve the accuracy of SNNs on real-world images. Spyker implements an efficient version of ZCA whitening by taking advantage of routines from highly optimized linear algebra libraries (BLAS and LAPACK) that operate on symmetric matrices. This layer is implemented as \cinline{spyker.ZCA} class which has a \cinline{fit(array, epsilon)} and a call function.

\subsection{Neural Coding}

SNNs process spike trains, but the input consists of analog values (for example, images are made of pixel values). In order to make these inputs suitable for the network, a conversion scheme is needed. The mapping from stimuli to neural responses is called neural coding \cite{panzeri_cracking_2017}. Coding schemes implemented in Spyker are explained in the following subsections.

\subsubsection{Rate Coding}

Out of several coding schemes suggested, rate coding is widely used where the rate of firing of the neurons represents information. In this scheme, the rate of firing is dependent on the intensity of the input value (higher intensity corresponds to faster firing) \cite{adrian_impulses_1926}. The exact time of firing in each neuron is stochastic in nature and may be modeled with a Poisson distribution. A lengthy window of time is required to transmit the information in this coding, and the spikes are not quite sparse.

\subsubsection{Temporal Coding}

Another popular coding scheme is temporal coding \cite{thorpe_spike-based_2001}. Recordings in the primary visual cortex show \cite{masquelier_unsupervised_2007} that the response latency decreases with the stimulus contrast. This coding scheme can convey information through the timings of the spikes. Multiple forms of this scheme have been proposed, including rank order coding \cite{thorpe_rank_1998}. Instead of computing the exact timing of each spike, the timings are computed relative to one another in rank order coding. This relative (instead of exact) timing can increase invariance to changes in the input intensity and contrast \cite{thorpe_rank_1998}. It has been suggested \cite{gautrais_rate_1998} that temporal coding might be more efficient in some situations.

\subsubsection{Coding in Spyker}

Spyker supports rank order and rate coding. The concept of time is implemented with spikes occuring in time steps in this library. Rank order coding maps higher intensities to earlier time steps of a neuron firing. In order to calculate the time step the neuron will fire in, Spyker sorts the intensity values by default. This calculates rank order between spikes, and the spikes will be distributed among time steps evenly. The sorting operation is computationally expensive (specially on GPUs), and optionally, it can be disabled to have runtime improvements (however, accuracy might be affected). Since processing time steps sequantially is inefficient and time-consuming, Spyker processes all the time steps at once. To this end, when a neuron fires in time step $t_i$, it will also fire at time steps $t_{i+1}$, $t_{i+2}$, ..., $t_n$ where $n$ is the number of time steps. An example of this cumulative structure can be seen in Figure \ref{fig:filter-example}.

\subsection{Neural Model}

Once the input is filtered and coded, it gets processed by the network. The network is built using fully connected, convolution, integrate-and-fire (IF) activation, pooling, and padding layers. These operations are explained in the following subsections.

\subsubsection{Convolution}

The integrate-and-fire mechanism is implemented by combining convolution and the IF activation layer. The internal potentials of the neurons are computed using convolution operation, and the IF activation operation produces spikes where neurons have a potential higher than a specified threshold. Multiple layers can be assembled and stacked on top of one another to create deeper structures.

The convolution layer has a kernel with $C_o \times C_i \times K_h \times K_w$ shape. the synaptic weights are initialized randomly with a normal distribution. It performs two-dimensional convolution with support for padding and stride. The input has $B \times T \times C_i \times H_i \times W_i$ shape which corresponds to batch, time steps, channels, height, and width of the input, respectively. The output has $B \times T \times C_o \times H_o \times W_o$ shape where:
\begin{equation}
\begin{aligned}
H_o &= \lfloor \frac{H_i + 2 \times P_h - K_h}{S_h} \rfloor + 1 \\
W_o &= \lfloor \frac{W_i + 2 \times P_w - K_w}{S_w} \rfloor + 1
\end{aligned}
\end{equation}
And $P_h$, $P_w$, $S_h$, $S_w$ are the height and width of convolution padding and stride. Padding increases the size of the two-dimensional input before convolution operation by expanding the edges of the input and filling in the new space with a constant value (usually zero). Stride is the number of steps the convolution window takes when it moves on the image. The output of the convolution layers are internal potentials of neurons that need to be passed through an IF activation layer to become output spike trains. This layer is implemented with \cinline{spyker.Conv(insize, outsize, kernel, stride, pad, mean, std, device)} class in Spyker.

\subsubsection{Fully Connected}

The fully connected layer is combined with the IF activation to model the IF neurons, much similar to what happens in the convolution layers. This layer has a kernel with $I \times O$ shape. The synaptic weights are initialized randomly with a normal distribution. The input has $B \times T \times I$ which corresponds to batch, time steps, and input size, respectively. The output has $B \times T \times O$ shape. The fully connected layer is represeneted by \cinline{spyker.FC(insize, outsize, mean, std, device)} in the library.

\subsubsection{Pooling}

The pooling layer performs two-dimensional max pooling operation with a window size of$L_h \times L_w$, a stride of $S_h \times S_w$, and a padding of $P_h$, $P_w$. The input has $B \times T \times C_i \times H_i \times W_i$ shape and the output has $B \times T \times C_o \times H_o \times W_o$ shape where:
\begin{equation}
\begin{aligned}
H_o &= \lfloor \frac{H_i + 2 \times P_h - L_h}{S_h} \rfloor + 1 \\
W_o &= \lfloor \frac{W_i + 2 \times P_w - L_w}{S_w} \rfloor + 1
\end{aligned}
\end{equation}
The interface of Spyker has the \cinline{spyker.pool(array, kernel, stride, pad, rates)} function to run the pooling operation on the input given the kernel, stride, and padding size. \cinline{rates} argument is the rate of firing of the neurons when rate coding is used. The pooling operation selects neurons that fire earlier when rank order coding is used, and selects neurons that have a higher firing rate when rate coding is used.

\subsection{Learning}

Learning in the brain happens when the strength of connections change between its neurons, and this change in strength is named synaptic plasticity \cite{citri_synaptic_2008}. Learning methods that utilize synaptic plasticity have been developed for SNNs \cite{markram_regulation_1997, gerstner_neuronal_1996, fremaux_neuromodulated_2016}.

\subsubsection{Spike-timing-dependent Plasticity}

One widely recognized synaptic plasticity learning rule is spike-timing-dependent plasticity (STDP) \cite{markram_regulation_1997, gerstner_neuronal_1996}. STDP learning rule operates by adjusting synaptic weights and utilizing the timing of the spikes. A pre-synaptic neuron firing before (after) the post-synaptic neuron results in a strengthed (weakened) connection. STDP allows the neurons to extract and learn frequent features in the input \cite{bichler_extraction_2012}. STDP layer changes synaptic weights with stabilization:
\begin{equation}
\Delta W_{i,j} =
\begin{cases}
A^{+}_k (W_{i,j} - L_k) (U_k - W_{i,j}), & t_j \leq t_i \\
A^{-}_k (W_{i,j} - L_k) (U_k - W_{i,j}), & t_j \ge t_i \\
\end{cases}
\end{equation}
where $A^+_k$, $A^-_k$, $L_k$, $U_k$ are the positive learning rate, negative learning rate, lower bound, and upper bound of the $k$th configuration, respectively. If stabilization is not set, then the formula becomes:
\begin{equation}
\Delta W_{i,j} =
\begin{cases}
A^{+}_k, & t_j \leq t_i \\
A^{-}_k, & t_j \ge t_i \\
\end{cases}
\end{equation}
then the weights are computed:
\begin{equation}
W^+_{i,j} = max(L_k, min(U_k, \Delta W_{i,j}))
\end{equation}

Input, neurons selected by the winner-take-all mechanism (WTA), and the output are passed to a function belonging to fully connected or convolution layers, and the STDP learning rule is applied. Convolution or fully connected layers in Spyker can have multiple STDP configurations (different learning rules, weight clipping, enabling/disabling stabilizer) implemented as \cinline{spyker.STDPConfig(positive, negative, stabilize, lower, upper)}.  Each winner neuron can be mapped to an STDP configuration, and that neuron will be updated using the learning rates and such that belongs to the selected configuration. SpykeTorch creates an STDP object for each configuration, and mapping winner neurons to different configurations is done by the user. Compared to SpykeTorch, Spyker provides a more flexible and easy to use API for weight updating and enables batch updating, which improves performance. Samples are processed in mini-batches which increases performance drastically (see the results section), and the batch update rule does not differ from single-sample processing.

\subsubsection{Reward-modulated STDP}

Another approach is using the reinforcement (RL) learning rule. One method based on RL is reward-modulated STDP \cite{fremaux_neuromodulated_2016}. R-STDP adjusts the STDP such that neurons that respond correctly are rewarded, and punished otherwise. It has been suggested \cite{mozafari_bio-inspired_2019} that when the input has non-diagnostic frequent features that are less effective in decision-making, R-STDP is able to discard these features and improve the decision-making process. Since convolution and fully connected layers accept STDP configurations as input, R-STDP can be implemented by passing two configurations to a layer (one for rewarding and one for punishing), and mapping each winner neuron to a configuration based on data labels. If one formulates this, $\Delta W_{i,j}$ will be:
\begin{equation}
\begin{cases}
\begin{cases}
A^{+}_r (W_{i,j} - L_r) (U_r - W_{i,j}), & t_{pre} < t_{post} \\
A^{-}_r (W_{i,j} - L_r) (U_r - W_{i,j}), & t_{pre} \geq t_{post} \\
\end{cases}, & \text{if reward} \\ \\
\begin{cases}
A^{-}_p (W_{i,j} - L_p) (U_p - W_{i,j}), & t_{pre} < t_{post} \\
A^{+}_p (W_{i,j} - L_p) (U_p - W_{i,j}), & t_{pre} \geq t_{post} \\
\end{cases}, & \text{if punish} \\
\end{cases}
\end{equation}

\subsubsection{Winner-take-all and Lateral Inhibition}

When a neuron fires at a specific location, lateral inhibition \cite{goriely_determination_1991, heitzler_choice_1991} operation inhibits other neurons belonging to other neural maps from firing in that location. Lateral inhibition for the convolution operation can be used with \cinline{spyker.inhibit(array, thershold, inplace)} functions. Winner neurons that STDP weight updating will be performed on are selected by the winner-take-all \cite{maass_computational_2000, oster_computation_2009} operation. WTA selects neurons that fire earlier, and if the firing time of neurons is the same, then the one that has a higher internal potential will be selected. This operation is implemented with \cinline{spyker.fcwta(array, radius, count, threshold)} for fully connected and \cinline{spyker.convwta(array, radius, count, threshold)} for convolution operations.

%%%%%%%%%%%%%%%%%%%%%%%%%%%%%%%%%%%%%%%%%%%%%%%%%%%%%%%%%%%%%%%%%%%%%%%%%%%%%%%%
% RESULTS %%%%%%%%%%%%%%%%%%%%%%%%%%%%%%%%%%%%%%%%%%%%%%%%%%%%%%%%%%%%%%%%%%%%%%
%%%%%%%%%%%%%%%%%%%%%%%%%%%%%%%%%%%%%%%%%%%%%%%%%%%%%%%%%%%%%%%%%%%%%%%%%%%%%%%%

\section{Results}

In this section, we will test the performance of the library against the SpykeTorch library. Afterward, a comparison of the represented stimuli extracted from Spyker to recorded electrophysiology data is conducted to demonstrate the applicability of SNNs in describing the underlying neural mechanisms of brain functions.

\subsection{Library Performance}

In this section, we compare the performance of the library to SpykeTorch on two networks that classify the MNIST dataset.

\subsubsection{R-STDP Network}

\begin{figure}[tb]
\centerline{\includegraphics[width=\columnwidth]{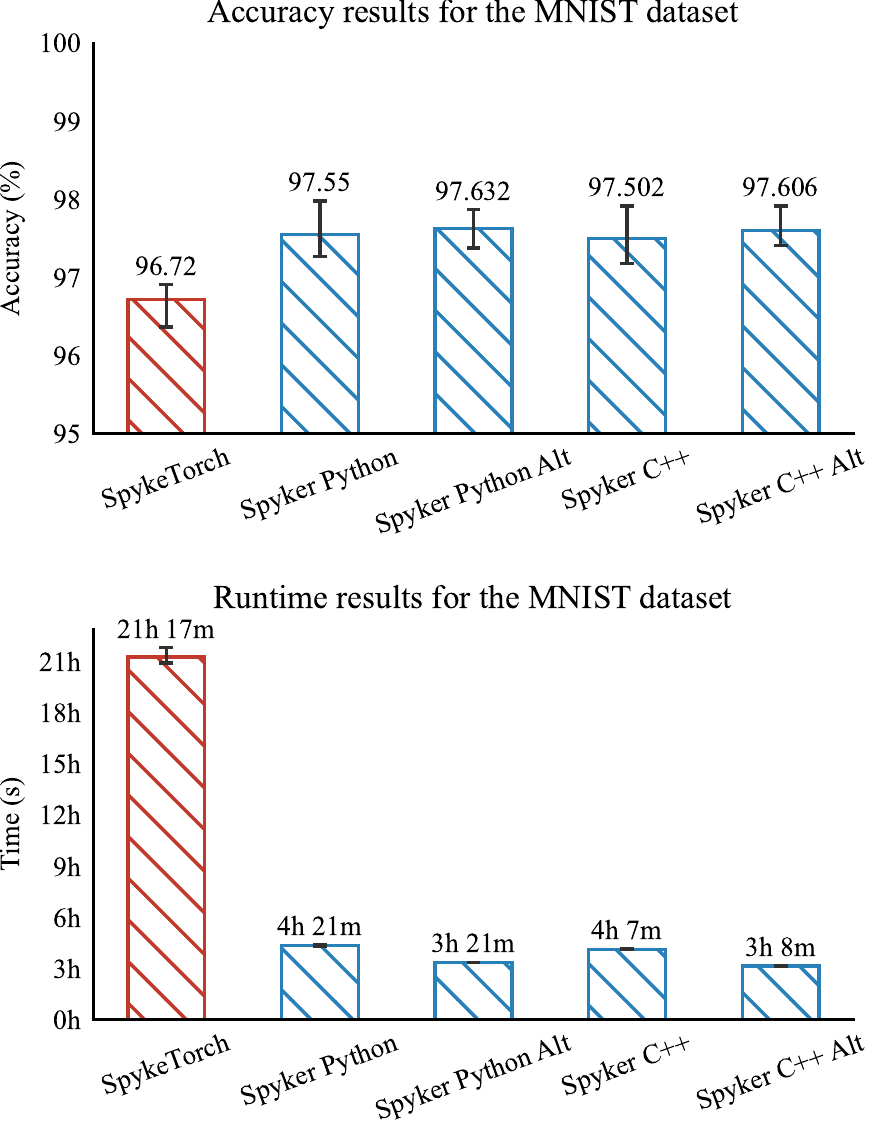}}
\caption{Comparison plots of the runtime and accuracy of Spyker aganist SpykeTorch on the Mozafari et al. network. The plot on the left shows the runtime comparison of Spyker and SpykeTorch implementations. The plot on the right also compares accuracy of the two implementations. Comparisons are between SpykeTorch (ST), implementation using Spyker in Python (SP Py), alternative version using Spyker in Python (SPA Py), and their C++ counterparts (SP C++, SPA C++). The error bars are minimum and maximum values of the samples.}
\label{fig:mozaf-results}
\end{figure}

The first netwrok is the Mozfari et al. network \cite{mozafari_bio-inspired_2019} which has three convolutional layers. The first layer is trained two times with STDP, the second layer four times with STDP, and the third layer 680 times with R-STDP on the training set while compuing the test accuracy at each iteration while training the third layer. We made a small change to the structure of the network (named Alt for alternative). We removed the input padding from the last convolution layer and changed its window size to 4 and the output channels to 400. Results can be seen in Figure \ref{fig:mozaf-results} and Table \ref{tab:mozaf-results}. All the tests are performed on Inte Core i7-9700k with 64G memory and Nvidia Geforce GTX 1080 Ti with 12G memory (Ubuntu 18.04).

\begin{table}[htbp]
\caption{Comparisons of the the runtime and accuracy of Spyker aganist SpykeTorch on the Mozafari et al. network.}
\label{tab:mozaf-results}

\begin{center}
\begin{tabular}{|c|c|c|c|c|}
\hline
Implementation    & Time   & \begin{tabular}[c]{@{}c@{}}Time\\ (S±SD)\end{tabular} & \begin{tabular}[c]{@{}c@{}}Accuracy\\ (\%±SD)\end{tabular} & Runs \\ \hline
SpykeTorch        & 21h17m & 76,672±916                                            & 96.720±0.163                                               & 12   \\ \hline
Spyker Python     & 04h49m & 15668±52                                              & 97.550±0.169                                               & 30   \\ \hline
Spyker Python Alt & 03h31m & 12,114±14                                             & 97.632±0.112                                               & 30   \\ \hline
Spyker C++        & 03h52m & 14,869±50                                             & 97.502±0.157                                               & 30   \\ \hline
\end{tabular}
\end{center}
\end{table}

In order to compare the results, we test whether the two-sample mean difference confidence interval (99.9\%) contains zero. The null hypothesis is having the same means, and the alternative is having different means. The test results indicate that the Spyker Python implementation is faster compared to the SpykeTorch implementation (Confidence intervals are [15477, 15859] and [72607, 80737] for Spyker and SpykeTorch respectively, showing no intersection). Furthermore, the alternative implementation is faster both in the Python implementation with [-3738, -3370] interval and the C++ implementation with [-3828, -3339] interval. As expected, the C++ interface is faster compared to the Python interface with [-1078, -520] interval. The results for the accuracy comparisons show that there are no significant differences ([96.932, 98.169] and [95.996, 97.444] for Python vs SpykeTorch implementations respectively, showing intersection, [-0.89, 0.793] for C++ vs Python, [-0.649, 0.813] for Python alternative vs Python, and [-0.763, 0.971] for C++ alternative vs C++).

\subsubsection{STDP Network}

\begin{figure}[tb]
\centerline{\includegraphics[width=\columnwidth]{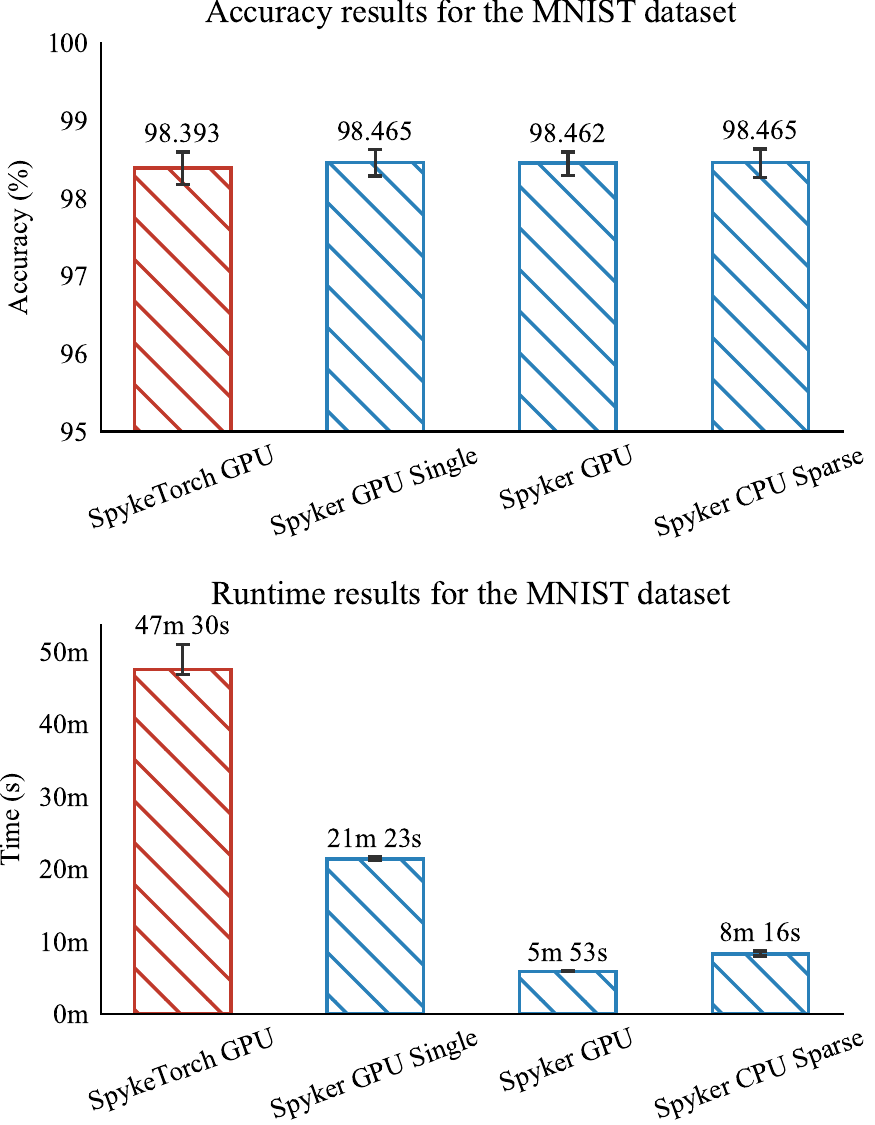}}
\caption{Comparison plots of the runtime and accuracy of Spyker against SpykeTorch on the Kheradpisheh et al. network. The plot on the left shows shows the runtime comparison of Spyker and SpykeTorch implementations. The plot on the right also compares accuracy of the two implementations. Comparisons are between GPU implementation using SpykeTorch (SP GPU), GPU implementation using Spyker with single-sample instead of batch processing (SP Single), GPU implementation using Spyker (SP GPU), and Sparse CPU implementation using Spyker (SP Sparse). The error bars are minimum and maximum values of the samples.}
\label{fig:kherad-results}
\end{figure}

Subsequently, the Kheradpisheh et al. network \cite{kheradpisheh_stdp-based_2018} is used for comparisons. This network is made of two convolutional layers. The first layer is trained 2 times with STDP, and the second layer is trained 20 times with STDP on the training set. The output of the network is classified uing the SVM classifier. The elapsed time measured consists of the time needed to train the network on the training set and make predictions for the testing set. The time to utilize SVM is not taken into account because the libraries that simulate the neural network portion are compared here. The results can be seen in in Figure \ref{fig:kherad-results} and Table \ref{tab:kherad-results}.

\begin{table}[htbp]
\caption{Comparisons of the the runtime and accuracy of Spyker aganist SpykeTorch on the Kheradpisheh et al. network.}
\label{tab:kherad-results}

\begin{center}
\begin{tabular}{|c|c|c|c|c|}
\hline
Implementation    & Time   & \begin{tabular}[c]{@{}c@{}}Time\\ (S±SD)\end{tabular} & \begin{tabular}[c]{@{}c@{}}Accuracy\\ (\%±SD)\end{tabular} & Runs \\ \hline
SpykeTorch GPU    & 47m30s & 2,850±64                                              & 98.392±0.093                                               & 30   \\ \hline
Spyker GPU Single & 21m23s & 1,283±6                                               & 98.465±0.095                                               & 30   \\ \hline
Spyker GPU        & 05m53s & 353±9                                                 & 98.461±0.079                                               & 30   \\ \hline
Spyker Sparse     & 08m16s & 496±1                                                 & 98.464±0.091                                               & 30   \\ \hline
\end{tabular}
\end{center}
\end{table}

The test results indicate that the Spyker GPU implementation is faster compared to the SpykeTorch implementation (confidence interval [-2728, -2265]). Since the SpykeTorch implementation processes one sample at a time, we also implemented a single sample version on the GPU, and this implementation runs faster compared to the SpykeTorch implementation (confidence interval [-1795, -1338]). There is also an implementation using the sparse interface of the Spyker (that runs on CPU) that is faster than the SpykeTorch implementation on the GPU (confidence interval [-2586, -2120]). These results show that the Spyker implementation is faster while the accuracy is not significantly different ([-0.373, 0.511] for Spyker GPU, [-0.458, 0.603] for single-sample, and [-0.405, 0.549] for sparse implementation, all against the SpykeTorch implementation).

\subsection{Analyzing the Underlying Structures of the Brain}

In order to demonstrate the use case and the importance of the library in neuroscience research, a similarity analysis is done in this section to compare the biological plausibility of an SNN and a deep CNN model. The neural data needed for the analysis is recorded as spiking activity and LFP signals from Inferior Temporal (IT) cortex using a single electrode (169 sessions from two macaque monkeys, the neural data for the monkeys are pooled together) \cite{farhang_effect_2021}. The task implemented here is a Rapid Serial Visual Presentation (RSVP). The intervals are 50ms for stimulus and 450ms interstimulus. Eighy-one greyscale images of real-world objects and Gaussian low-pass filtered and high-pass filtered variations of some are shown during the task (total 155 images). The categories of the stimuli are animal face (AF), human face (HF), animal body part(AB), human body part (HB), natual objects (N), and man-made objects (MM).

The SNN used here is structurally similar to the one introduced by Shirsavar et al. \cite{shirsavar_faster_2022}. The input of the SNN is resized to 32 and passed through 3 LoG fitlers with stds of 0.471, 1.099, 2.042. The window sizes of the filters are 7. Then, the output is thresholded and coded into 15 time steps. The first convolution layer has 16 output channels with awindow size of 5 and a padding of 2, and the second convolution layer has 32 output channels with a window size of 3 and a padding of 1. The pooling layers have 2 and 3 window sizes, respectively. The layers are trained 20 times on the images, and the learning rates are doubled after each image until they reach 0.15. Firing times (divided by number of time steps) of the final layer is used as the network output.

\begin{figure*}
\includegraphics[width=\textwidth]{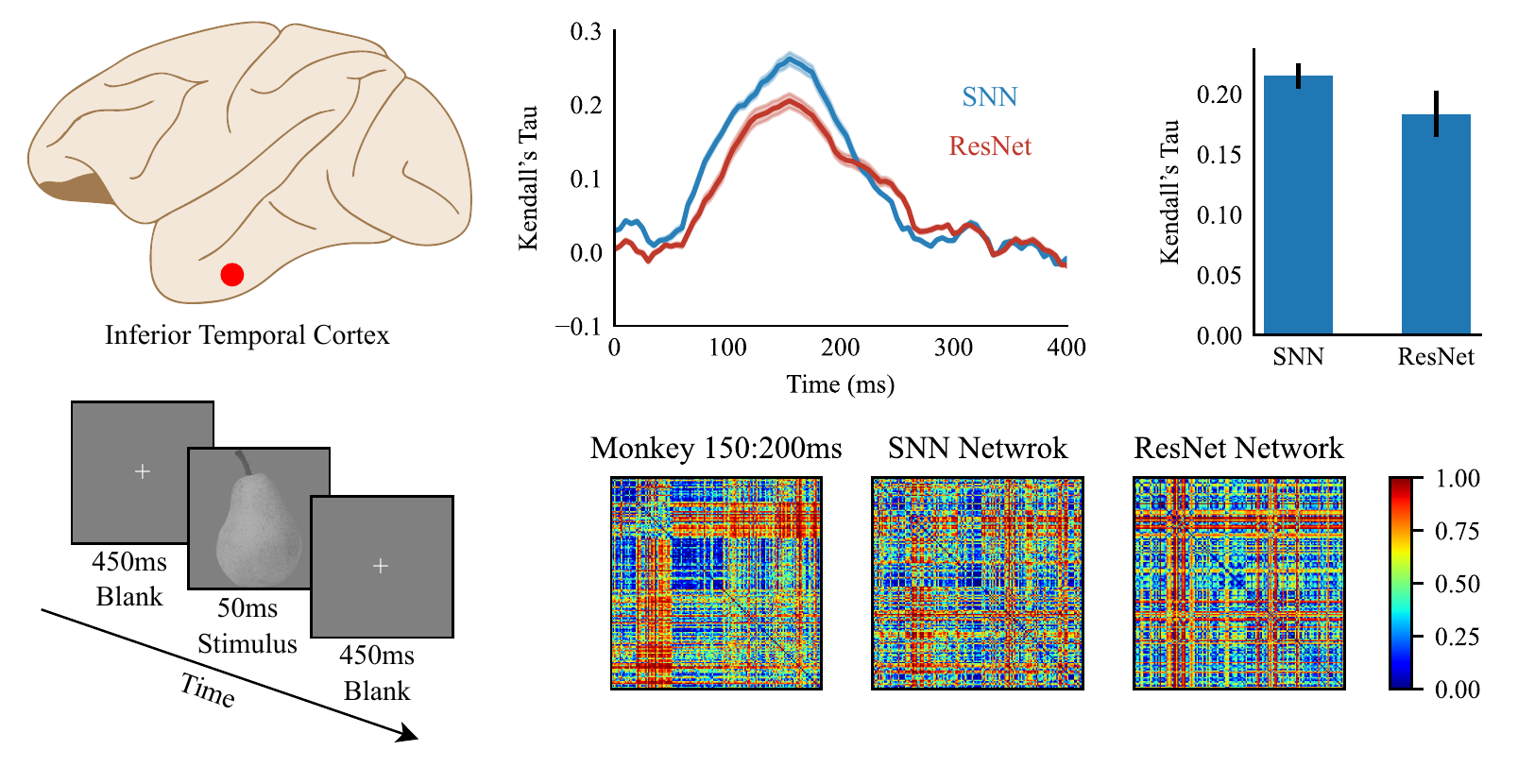}
\caption{Similarity comparison of SNN and ResNet-50 to monkey neural data. The similarity measurement used here is the cosine similarity. The RDM for the monkey is computed for the 50ms interval after the onset. The RDMs are adjusted with histogram equalization. The RSA is calculated with 50ms window size and 5ms stride and 95\% confidence interval. Kendall's Tau measurement is used for the RSA analysis. The RSA is averaged in the interval between 125ms and 175ms and compared in the plot in the top right with 95\% confidence interval.}
\label{fig:rsa-plot}
\end{figure*}

The CNN network used here is a ResNet-50 with the classifier layer replaced. The network is not pretrained. The input image is resized to 256 and cropped to 224. The network is trained 15 times on the dataset with Adam optimizer and 0.0001 learning rate. using a linear SVM classifier to classify the 6 categories. the accuracies for the 6 classes are 51.569 $\pm$ 2.240 (SD), 48.623 $\pm$ 2.538, and 51.247 $\pm$ 2.257 for ResNet-50, SNN, and an SVM classifier that is used on the average firing rates of the neural recordings of the monkeys between 150ms and 200m from the onset, respectively. Figure \ref{fig:rsa-plot} Shows the results of the analysis. The average Kendall's Tau value for the interval between 125ms and 175ms shown in the figure is tested between the SNN and the ResNet. Using a Mann-Whitney U test with the alpha value of 0.001 results in a p-value of 2.028-07, which shows significant difference between the two. This indicates that the SNN has a closer structure to monkey brain.

\subsection{Rate Coding Output}

\begin{figure*}
\includegraphics[width=\textwidth]{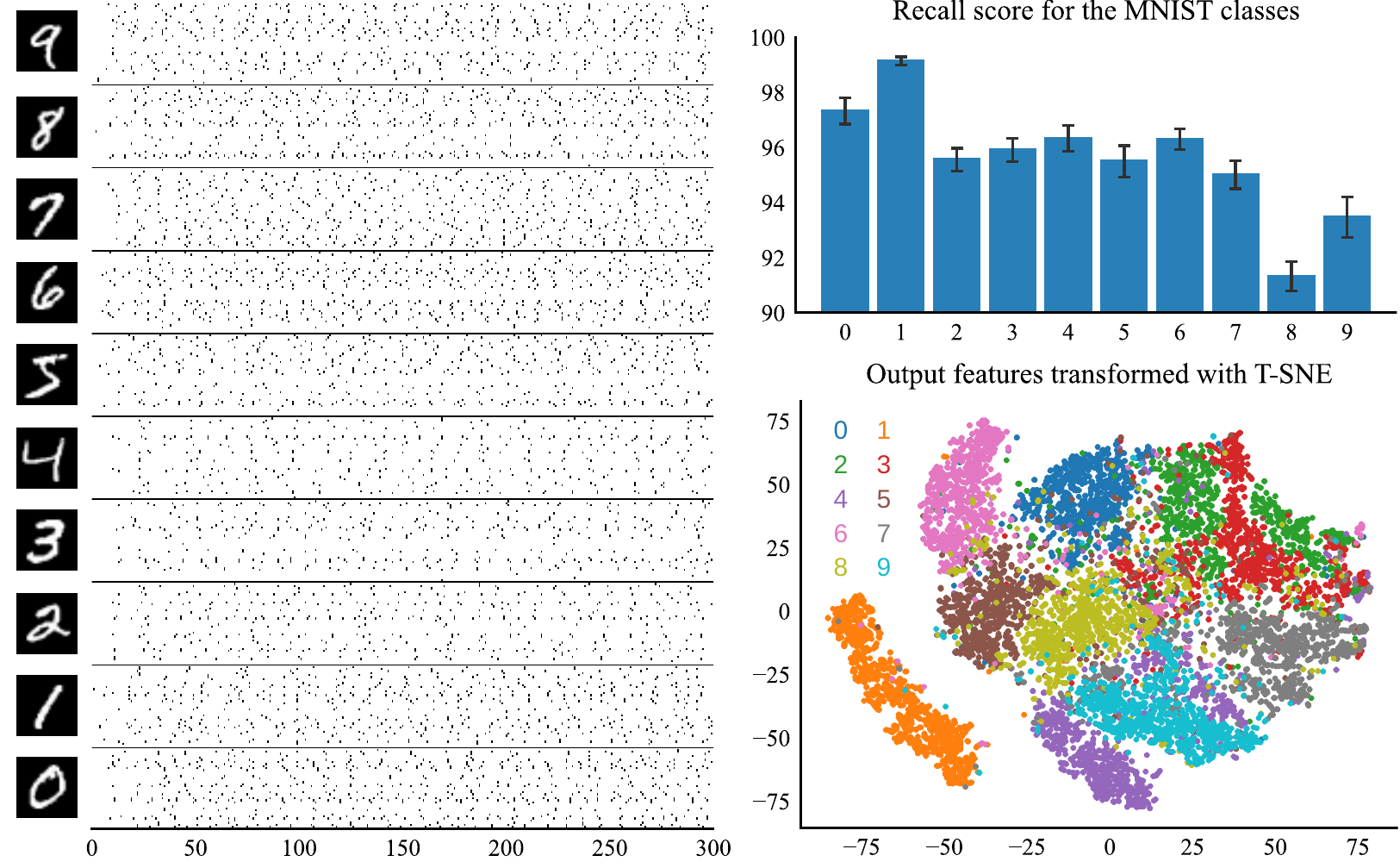}
\caption{Raster plot of an SNN network for the MNIST test images. In this figure, 40 neurons are plotted in 300 time steps for 10 samples of the MNIST testing set, each image belonging to one class.}
\label{fig:raster-plot}
\end{figure*}

In this section, we look at the output of an SNN that uses rate coding. The SNN network used here is the Shirsavar et al. \cite{shirsavar_faster_2022}. The number of output channels in the convolutional layers are set to 25 and 50. The training is not changed in that 15 time steps are used with rank order coding. However, the inference is done with 300 time steps and rate coding. Afterward, the spike output of 40 neurons are plotted for one testing sample for each class shown in Figure \ref{fig:raster-plot}. The figure also cointains a plot of T-SNE transformed firing rates as output fetures and the recall score for each class for the average of 30 runs. The accuracy of the 30 runs is 95.635$\pm$0.171 on the testing set.

%%%%%%%%%%%%%%%%%%%%%%%%%%%%%%%%%%%%%%%%%%%%%%%%%%%%%%%%%%%%%%%%%%%%%%%%%%%%%%%%
% API DEMONSTRATOIN %%%%%%%%%%%%%%%%%%%%%%%%%%%%%%%%%%%%%%%%%%%%%%%%%%%%%%%%%%%%
%%%%%%%%%%%%%%%%%%%%%%%%%%%%%%%%%%%%%%%%%%%%%%%%%%%%%%%%%%%%%%%%%%%%%%%%%%%%%%%%

\section{Library Demonstration}

In this section, a sample usage of the library is illustrated. The network used here is introduced by Shirsavar et al. \cite{shirsavar_faster_2022} to classify the MNIST dataset. The network has two convolutional layers trained with the STDP learning rule. The code shown in this section is only a part of the actual implementation, with the aim of providing a simple example. For the complete implementation, please visit the GitHub repository of Spyker\footnote{\url{https://github.com/ShahriarRezghi/Spyker}}.

\subsection{Transformation}

The transformation from the input image to the network input consists of feature enhancement and spike coding, shown in Listing \ref{lst:transform-code}. Here, a module named Transform is defined that performs the transformation when called. This module applies 3 LoG filters with different standard deviations to the input image with padding to keep the original width and height of the input. The output is stored in 6 channels. Each channel of this output is then coded into fifteen time steps using rank order coding.

\begin{lstlisting}[caption={Implementation of the Transform module}, label={lst:transform-code}]
class Transform:
  def __init__(self, device):
    std = [0.471, 1.099, 2.042]
    self.filt = spyker.LoG(3, std,
      pad=3, device=device)

  def __call__(self, data):
    data = self.filt(data)
    spyker.threshold(data, 0.01)
    return spyker.code(data, 15)
\end{lstlisting}

\subsection{Network}

The network has two convolutional layers. Here, a module named Network is defined (shown in Listing \ref{lst:network-code}) to train the neurons and make predictions. Here, the convolution layers are initialized, STDP configurations are set, and the winner selection function is wrapped with a lambda function to keep the hyperparameters in the initialization of the function of the network.

\begin{lstlisting}[caption={Implementation of the Network module}, label={lst:network-code}]
class Network:
  def __init__(self, device):
    self.thresh1, self.thresh2 = 16, 5
    self.conv1 = spyker.Conv(6, 100, 5,
      pad=2, mean=.5, std=.02, device=device)
    self.conv2 = spyker.Conv(100, 200, 3,
      pad=1, mean=.5, std=.02, device=device)
    config1 = spyker.STDPConfig(.0004, -.0003)
    config2 = spyker.STDPConfig(.0004, -.0003)
    self.conv1.stdpconfig = [config1]
    self.conv2.stdpconfig = [config2]
    self.wta1 = lambda x: spyker.convwta(x, 3, 5)
    self.wta2 = lambda x: spyker.convwta(x, 1, 8)
\end{lstlisting}

\subsection{Learning}

Training each layer is done in a separate function shown in Listing \ref{lst:intrain-code}. The training of the layers is done in a sequantial order (one layer after another). Training of the first layer is done in the train\_layer1 function with the STDP learning rule. Here, the output of the first convolution is computed, and lateral inhibition is performed on it. Then, winner neurons are selected, and STDP weight updating is performed on them. The STDP learning rates in the first layer are multiplied by 1.5 every 2000 samples, and the multiplying process stops once the positive learning rate reaches 0.15. The second layer is trained in a similar way in the train\_layer2 function with the STDP learning rule.

\begin{lstlisting}[caption={The code for training of the network layers}, label={lst:intrain-code}]
def train_layer1(self, data):
  output = self.conv1(data)
  spyker.threshold(output, self.thresh1)
  spyker.inhibit(output)
  winners = self.wta1(output)
  spikes = spyker.fire(output)
  self.conv1.stdp(data, winners, spikes)

def train_layer2(self, data):
  data = self.conv1(data)
  data = spyker.fire(data, self.thresh1)
  data = spyker.pool(data, 2)
  output = self.conv2(data)
  spyker.threshold(output, self.thresh2)
  spyker.inhibit(output)
  winners = self.wta2(output)
  spikes = spyker.fire(output)
  self.conv2.stdp(data, winners, spikes)
\end{lstlisting}

After defining the network module, the process of training and classification is implemented. The training process shown in Listing \ref{lst:outrain-code} involves training each layer once with quantization afterward.

\begin{lstlisting}[caption={The training process of the network}, label={lst:outrain-code}]
for data, target in trainset:
  network.train_layer1(transform(data))
spyker.quantize(network.conv1.kernel, 0, 0.5, 1)

for data, target in trainset:
  network.train_layer2(transform(data))
spyker.quantize(network.conv2.kernel, 0, 0.5, 1)
\end{lstlisting}

\subsection{Inference}

The call operator of the network shown in Listing \ref{lst:call-code} implements the prediction procedure which processes the input spikes and produces the final network output.

\begin{lstlisting}[caption={Inference function of the network}, label={lst:call-code}]
def __call__(self, data):
  data = self.conv1(data)
  data = spyker.fire(data, self.thresh1)
  data = spyker.pool(data, 2)
  data = self.conv2(data)
  data = spyker.fire(data, self.thresh2)
  data = spyker.pool(data, 3)
  return spyker.gather(data).flatten(1)
\end{lstlisting}

After training, the output features for every sample in the training set and the testing set are computed (in the \cinline{gather} function). Then, an SVM classifier is trained on the training set outputs. Finally, predictions are made for the testing set outputs (shown in Listing \ref{lst:class-code}).

\begin{lstlisting}[columns=flexible, caption={Implementation of the dimension reduction and classification operations}, label={lst:class-code}]
xtr, ytr = gather(network, transform, train)
xte, yte = gather(network, transform, test)
svm = LinearSVC(C=2.4).fit(xtr, ytr)
pred = svm.predict(xte)
accuracy = (pred == testy.numpy()).mean()
\end{lstlisting}

%%%%%%%%%%%%%%%%%%%%%%%%%%%%%%%%%%%%%%%%%%%%%%%%%%%%%%%%%%%%%%%%%%%%%%%%%%%%%%%%
% DISCUSSION %%%%%%%%%%%%%%%%%%%%%%%%%%%%%%%%%%%%%%%%%%%%%%%%%%%%%%%%%%%%%%%%%%%
%%%%%%%%%%%%%%%%%%%%%%%%%%%%%%%%%%%%%%%%%%%%%%%%%%%%%%%%%%%%%%%%%%%%%%%%%%%%%%%%

\section{discussion}

Our brain has amazing capabilities. It can learn and perform complicated tasks in a robust manner and with low power consumption. Artificial neural networks have been created to mimic the power of the brain processes. Deep neural networks are ANNs that have had major success in recent years. However, there are structural differences between these networks and the brain, and they encounter problems when it comes to tolerance, energy, and sample efficiency. Spiking neural networks are the next generation of artificial neural networks. SNNs are not a new concept. However, they have been brought to attention recently due to their promising characteristics. The aim of these networks is to build a better model of the brain compared to DNNs.

Several well-established simulation tools exist for DNNs. These tools have allowed DNNs to reach their great success faster and have helped them to computationally scale up. SNNs lack such high-performance simulation tools. There have been some attempts at creating such tools, but they have not been able to live up to expectations. In this work, we introduced Spyker, a high-performance library written from scratch using low-level tools to simulate spiking neural networks on both CPUs and GPUs. Despite being stand-alone, Spyker has great flexibility and the ability to integrate with other tools to create a smooth developing experience. We compared the performance of this library with SpykeTorch, a simulation tool built on the PyTorch framework. We showed that Spyker is multiple times faster compared to this library. Furthermore, to demonstrate the applicability of SNNs in describing the underlying neural mechanisms of the brain functions and the role of Spyker in this field, we compared the similarity of a spiking neural network implemented with this library with the similarity of the ResNet model to a macaque monkey brain. Finally, we illustrated an example implementation to demonstrate the easy and modern interface of the library.

Strong SNN models can be implemented using the Spyker library to solve real-world machine learning problems. Features like fast processing and having a C++ interface alongside the Python interface make this library ready for both research and production. Generalization is an important concept in machine learning and having neural networks that learn and run fast are quite desirable. SNNs have the potential to become state-of-the-art models in machine learning. Other potential use cases of the library is to study and understand how the brain processes information using simulations. In other words, this library enables us to look at neuroscience through the eyes of a brain-inspired neural network.

Although this library has been shown to be performant, there is room for more improvements. Spyker has a sparse interface that runs on the CPU. The sparse interface can be extended to also run on the GPU, and this can improve the performance even further. Furthermore, the support for a larger number of neural models, coding schemes, and learning rules can be added. This helps the library to cover a great range of SNN building blocks. When choosing a model to be deployed on embedded and neuromorphic processors, SNNs are among the top choices due to their energy efficiency. SNNs are often used in neuromorphic computing. Another direction that Spyker can take is in this direction. The computational efficiency of the sparse interface of Spyker can be further improved and made compatible with these types of processors.

\end{document}